\documentclass[10pt,twocolumn,letterpaper]{article}

\usepackage{iccv}
\usepackage{times}
\usepackage{epsfig}
\usepackage{graphicx}
\usepackage{amsmath}
\usepackage{amssymb}
\usepackage[lined,boxed,ruled]{algorithm2e}
\usepackage{pifont}

\newcommand{\comment}[1]{}

\usepackage[pagebackref=true,breaklinks=true,letterpaper=true,colorlinks,bookmarks=false]{hyperref}

\iccvfinalcopy 

\def\iccvPaperID{119} 

\ificcvfinal\fi
\begin{document}

\title{Towards 3D Human Pose Estimation in the Wild: a Weakly-supervised Approach}
\comment{
\author{XingyiZhou\\
School of Computer Science\\
Fudan University, Shanghai, China\\
{\tt\small zhouxy13@fudan.edu.cn}
\and
Qixing Huang\\
The University of Texas at Austin\\
Austin, Texas, U.S.A. \\
{\tt\small huangqx@cs.utexas.edu}
\and
Xiao Sun\\
Microsoft Research\\
Beijing, China \\
{\tt\small xias@microsoft.com}
\and
Xiangyang Xue\\
School of Computer Science\\
Fudan University, Shanghai, China\\
{\tt\small xyxue@fudan.edu.cn}
\and
Yichen Wei\\
Microsoft Research\\
Beijing, China \\
{\tt\small yichenw@microsoft.com}
}
}

{
\author{Xingyi Zhou$^{1,2}$, Qixing Huang$^2$, Xiao Sun$^3$,  Xiangyang Xue$^1$, Yichen Wei$^3$ \\
$^1$Shanghai Key Laboratory of Intelligent Information Processing \\
School of Computer Science, Fudan University \\
	$^2$ The University of Texas at Austin\\
	$^3$ Microsoft Research\\
	{\tt\small  \{zhouxy13,xyxue\}@fudan.edu.cn, huangqx@cs.utexas.edu, \{xias, yichenw\}@microsoft.com}
}
}

\maketitle

\begin{abstract}
In this paper, we study the task of 3D human pose estimation in the wild. This task is challenging due to lack of training data, as existing datasets are either in the wild images with 2D pose or in the lab images with 3D pose. 

We propose a weakly-supervised transfer learning method that uses mixed 2D and 3D labels in a unified deep neutral network that presents two-stage cascaded structure. Our network augments a state-of-the-art 2D pose estimation sub-network with a 3D depth regression sub-network. Unlike previous two stage approaches that train the two sub-networks sequentially and separately, our training is end-to-end and fully exploits the correlation between the 2D pose and depth estimation  sub-tasks. The deep features are better learnt through shared representations. In doing so, the 3D pose labels in controlled lab environments are transferred to in the wild images. In addition, we introduce a 3D geometric constraint to regularize the 3D pose prediction, which is effective in the absence of ground truth depth labels. Our method achieves competitive results on both 2D and 3D benchmarks.
\end{abstract}


\section{Introduction}

\begin{figure}[t]
\begin{center}
 \includegraphics[width=0.95\linewidth]{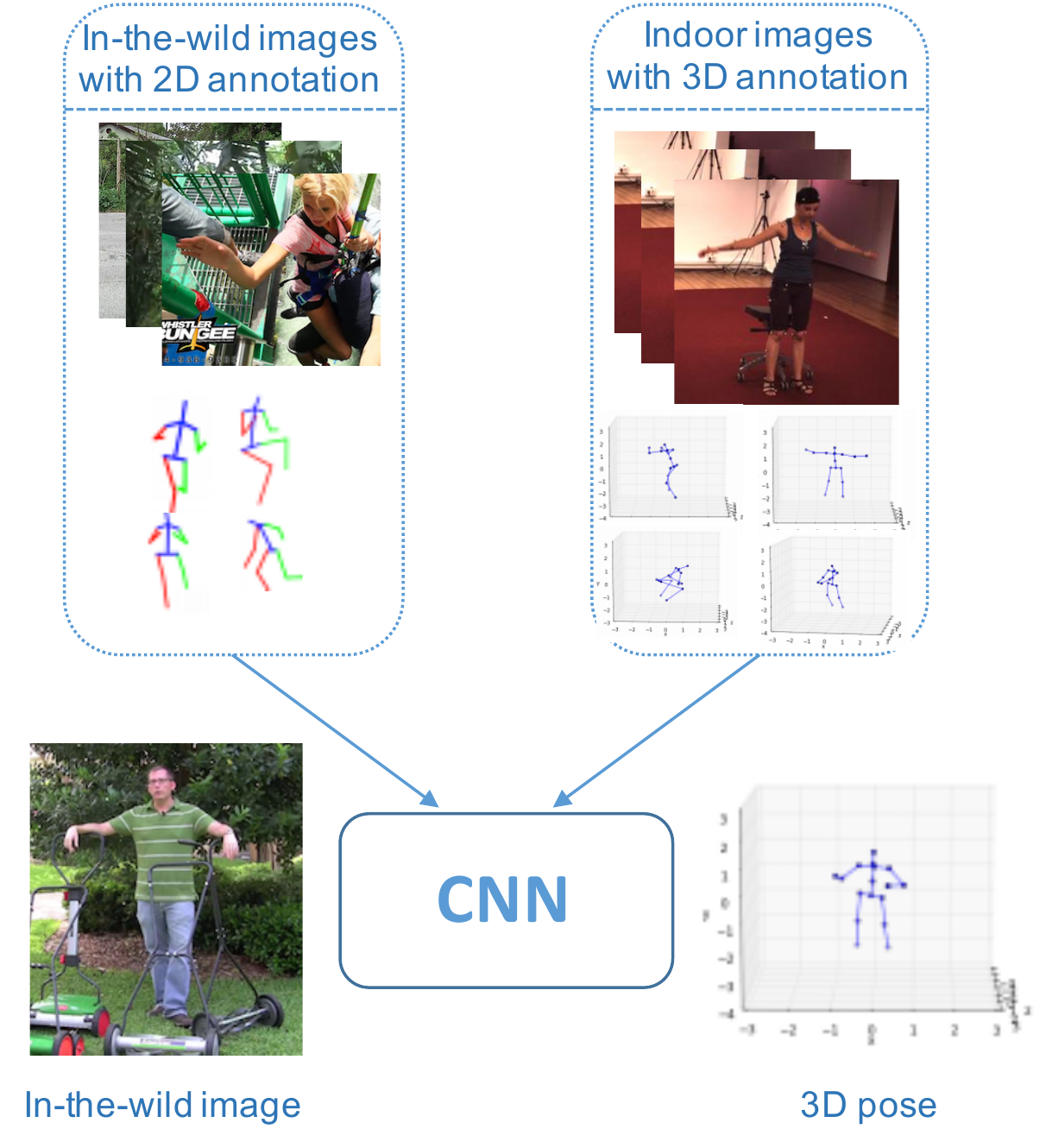}
\end{center}
   \caption{A schematic illustration of our method: transferring 3D annotation from indoor images to in-the-wild images. Top (Training):  Both indoor images with 3D annotation (Right) and in-the-wild images with 2D annotation (Left) are used to train the deep neural network. Bottom (Testing): The learned network can predict the 3D pose of the human in in-the-wild images.}
\label{fig:motivation}
\end{figure}

Human pose estimation problem has been heavily studied in computer vision. It has numerous important applications in human-computer interaction, virtual reality, and action recognition. Existing research works falls into two categories: 2D pose estimation and 3D pose estimation. Thanks to the availability of large-scale 2D annotated human poses and the emergence of deep neural networks, the 2D human pose estimation problem has gained tremendous success recently~\cite{newell2016stacked,wei2016convolutional,insafutdinov2016deepercut,bulat2016human,chu2017multi}. State-of-the-art techniques are able to achieve accurate predictions across a wide range of settings (e.g., on images in the wild~\cite{andriluka14cvpr}).

In contrast, advance in 3D human pose estimation remains limited. This is partially due to the ambiguity of recovering 3D information from single images, and partially due to the lack of large scale 3D pose annotation dataset. Specifically, there is not yet a comprehensive 3D human pose dataset for images in the wild. The commonly used 3D datasets~\cite{h36m_pami,sigal2010humaneva} were captured by mocap systems in controlled lab environments. Deep neural networks~\cite{li20143d,zhou2016deep} trained on these datasets do not generalize well to other environments, such as in the wild.

There has been quite a few works on 3D human pose estimation in the wild. They usually proceed in two sequential steps~\cite{zhou2016sparseness,tome2017lifting,chen20163d,bogo2016keep,wu2016single,yasin2016dual}. The first step estimates 2D joint locations~\cite{newell2016stacked,wei2016convolutional,insafutdinov2016deepercut}. The second step recovers a 3D pose from these 2D joints~\cite{ramakrishna2012reconstructing,zhou20153d,akhter2015pose}. Training in the two steps are performed separately. Namely, 2D pose predictions are trained from 2D annotations in the wild, and 3D pose recovery from 2D joints is trained from existing 3D MoCap data. 
Such a sequential pipeline is clearly sub-optimal because the original in-the-wild 2D image information, which contains rich cues for 3D pose recovery, is discarded in the second step.

Recently, Mehta et al.~\cite{mehta2016monocular} have shown that 2D-to-3D knowledge transfer, i.e., using pre-trained 2D pose networks to initialize the 3D pose regression networks can significantly improve 3D pose estimation performance. This indicates that the 2D and 3D pose estimation tasks are inherently entangled and could share common representations.

Inspired by this work, we argue that the inverse knowledge transfer, i.e., from 3D annotations of indoor images to in-the-wild images, offers an effective solution for 3D pose prediction in the wild. In this work, we introduce a unified framework that can exploit 2D annotations of in-the-wild images as weak labels for the 3D pose estimation task. In other words, we consider a weakly-supervised transfer learning problem, where the source domain consists of fully annotated images in restricted indoor environment and the target domain consists of weakly-labeled images in the wild.

Similar to previous works~\cite{zhou2016sparseness,tome2017lifting,chen20163d,bogo2016keep,wu2016single,yasin2016dual}, our network also consists of a 2D module and a 3D module. However, instead of merely feeding the output of the 2D module as input to the 3D module, our approach connect the 3D module with the intermediate layers of the 2D module. This allows us to share the common representations between the 2D and the 3D tasks. The network is trained end-to-end with both 2D and 3D data simultaneously. This distinguishes our work from all existing works.

To better regularize the learning of weakly-supervised 3D pose estimation, we introduce a geometric constraint for training the 3D module. The geometric constraint is based on the fact that relative bone length in a human skeleton remains approximately fixed. The effectiveness of this constraint is experimentally verified when adapting the 3D pose information from labeled images in indoor environments to unlabeled images in the wild.

This work makes the following contributions:

\begin{itemize}
\item For the first time, we propose an end-to-end 3D human pose estimation framework for in-the-wild images. It achieves state-of-the-art performance on several benchmarks. 

\item We propose a 3D geometric constraint for 3D pose estimation from images with only 2D joint annotations. It has low cost in memory and computation. It improves the geometric validity of estimated poses.

\end{itemize}
Code is publicly available at \url{https://github.com/xingyizhou/pose-hg-3d}.

\section{Related Work}

Human pose estimation has been studied considerably in the past~\cite{moeslund2001survey,sarafianos20163d}, and it is beyond the scope of this paper to provide a complete overview of the literature.
In this section, we focus on previous works on 3D human pose estimation, which are most relevant to the context of this paper. 
We will also discuss related works on imposing weakly-/un-supervised constraints for training neural networks.

\textbf{3D Human Pose Estimation.} 
Given well labeled data (e.g., 3D joint locations of a human skeleton~\cite{h36m_pami,sigal2010humaneva}), 3D human pose estimation can be formulated as a standard supervised learning problem. 
A popular approach is to train a neural network to directly regress joint locations~\cite{li20143d}. 
Recently, people have generalized this approach in different directions. 
Zhou et al.~\cite{zhou2016deep} propose to explicitly enforce the bone-length constraints in the prediction, using a generative forward-kinematic layer;
Tekin et al. ~\cite{tekin2016structured} embed a pre-trained auto-encoder at the top of the network.
In contrast these works, Pavlakos et al introduce a 3D approach, which regresses a volumetric representation of 3D skeleton~\cite{pavlakos2016coarse}.
Despite the performance gain on standard 3D pose estimation benchmark datasets, the resulting networks do not generalize to images in the wild due to the domain difference between natural images and the specific capture environments utilized by these benchmark datasets.

\begin{figure*}[t]
\begin{center}
 \includegraphics[width=0.95\linewidth]{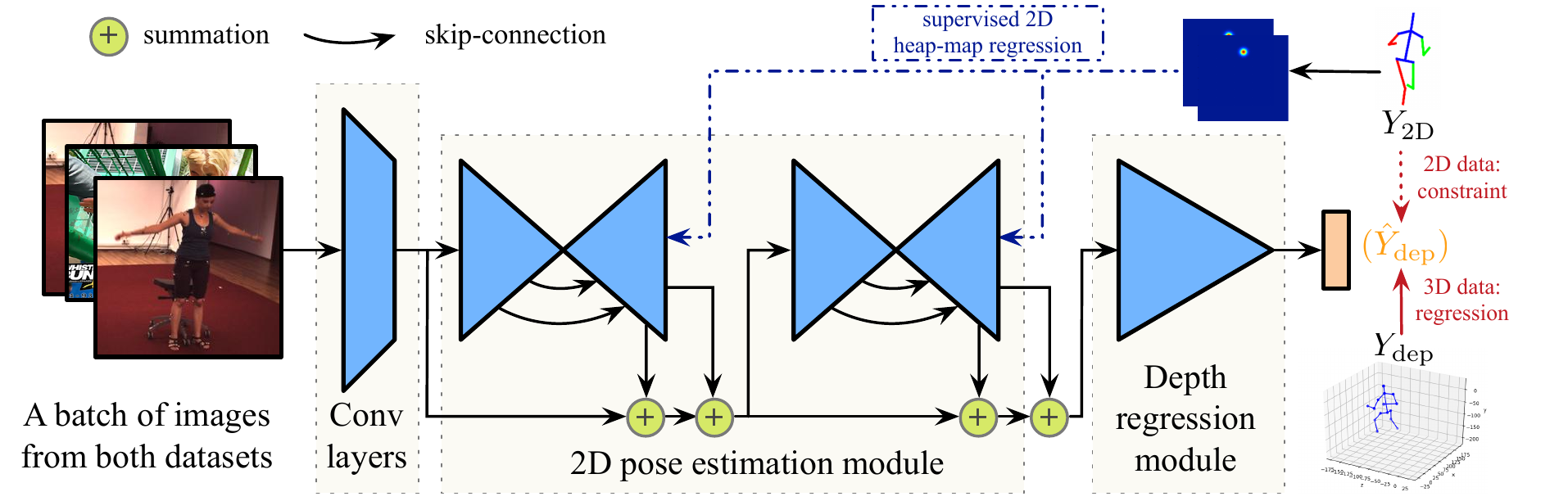}
\end{center}
   \caption{Illustration of our framework: In testing, images go through the stacked hourglass network and turn into 2D heat-maps. The 2D heat-maps and with lower-layer images features are summed as the input of the following depth regression module. In training, images from both 2D and 3D datasets are mixed in a single batch. For the 3D data, the standard regression with Euclidean Loss is applied. For the 2D data, we propose a weakly-supervised loss based on its 2D annotation and prior knowledge of human skeleton.}
\label{fig:Framework}
\end{figure*}

A standard approach to address the domain difference between 3D human pose estimation datasets and images in the wild is to split the task into two separate sub-tasks~\cite{zhou2016sparseness,tome2017lifting,chen20163d,bogo2016keep,wu2016single}. 
The first sub-task estimates 2D joint locations. This sub-task can utilize any existing 2D human pose estimation method (e.g., ~\cite{newell2016stacked,wei2016convolutional,insafutdinov2016deepercut,bulat2016human}) and can be trained from datasets of in-the-wild images. 
The second sub-task regresses the 3D locations of these 2D joints. 
Since the input at this step is just a set of 2D locations, the 3D pose estimation network can be trained on any benchmark datasets and then adapted in other settings.
Regarding 3D pose estimation from 2D joint locations, ~\cite{zhou2016sparseness} use an EM algorithm to compute a 3D skeleton by combining a sparse dictionary induced from the 2D heat-maps; 
~\cite{wu2016single,pavlakos2016coarse} use 3D pose data and its 2D projection to train a heatmap-to-3D pose network without the original image;
Bogo et al.~\cite{bogo2016keep} optimize both the pose and shape terms of a linear 3D human model~\cite{loper2015smpl} to best fit its 2D projection; 
Chen et al.~\cite{chen20163d} use nearest-neighbor search to match the estimated 2D pose to a 3D pose as well as a camera-view which may produce a similar 2D projection from a large 3D pose library; 
finally, Tome et al.~\cite{tome2017lifting} propose a pre-trained probabilistic 3D pose model layer that first generates plausible 3D human model from 2D heat-maps, and then refines these heat maps by combining 3D pose projection and image features.
All these methods, however, share a common limitation: the 3D pose is only estimated from the 2D joints, which is known to produce ambiguous results. 
In contrast, our approach leverages both 2D joint locations as well as intermediate feature representations from the original image.

An alternative approach for 3D human pose estimation is to train from synthetic datasets which are generated from deforming a human template model with known 3D ground truth~\cite{chen2016synthesizing,rogez2016mocap}. 
This is indeed a viable solution, but the fundamental challenge is how to model the 3D environment so that the distribution of the synthesized images matches that of the natural images. 
It turns out state-of-the-art methods along this line are less competitive on natural images.

There are also other works utilizing mixed 2D and 3D data for 3D human pose estimation. 
Mehta et al. ~\cite{mehta2016monocular} fine-tune a pre-trained 2D pose estimation network with 3D data. 
Popa et al. ~\cite{popa2017deep} consider 3D human pose estimation as a multi-task learning of 2D and depth regression with different data. 
Ours is different from those work that we use a weakly-supervised loss that seamlessly integrates both 2D and 3D data in a unified framework.

\textbf{Weakly-/un-supervised constraints.} 
In the presence of insufficient training data, incorporating generic or weakly supervised constraints among the prediction serves as a powerful tool for performance boosting. 
This idea was usually utilized in image classification or segmentation. 
Pathak et al.~\cite{pathak2015constrained} propose a constrained optimization framework that utilizes a linear constraint over sum of label probabilities for weakly supervised semantic segmentation. 
Tzeng et al.~\cite{tzeng2015simultaneous} propose a domain confusion loss to maximize the confusion between two datasets so as to encourage a domain-invariant feature. 
Recently, Hoffman et al.~\cite{hoffman2016fcns} introduce an adversarial learning based global domain alignment method and utilize a weak label constraint to apply fully connected networks in the wild. 
In this paper, we show this general concept can be used for pose estimation as well. 
To best of our knowledge, our approach is the first to leverage geometry-guided constraint to regularize the pose estimation network for images in the wild.

\section{Approach}

\subsection{Overview}

Given an RGB image $I$ containing a human subject, we aim to estimate the 3D human pose $Y \in \mathcal{Y}_{3D}$, represented by a set of 3D joint coordinates of the human skeleton,
i.e. $\mathcal{Y}_{3D} \in \mathcal{R} ^ {J \times 3}$, where $J$ is the number of joints.
We follow the convention of representing each 3D coordinate in the local camera coordinate system associated with $I$, namely, the first two coordinates are given by image pixel coordinates (which define the corresponding 2D joint location), and the third coordinate is the joint \emph{depth} in metric coordinates, \eg, millimeters in this work.

Our proposed network architecture is illustrated in Fig.~\ref{fig:Framework}. It consists of a 2D pose estimation module (Section ~\ref{sec:2D}) and a depth regression module (Section ~\ref{sec:lifting}). 
They predict the 2D joint locations $Y_{2D} \in \mathcal{Y}_{2D}$, where $\mathcal{Y}_{2D} \subset \mathcal{R}^{J\times 2}$, and the depth values $Y_{dep} \in  \mathcal{Y}_{dep}$, where $\mathcal{Y}_{dep} \subset \mathcal{R}^{J\times 1}$, respectively. The final output is the concatenation of $Y_{2D}$ and $Y_{dep}$.

The network is trained from both images in the lab with 3D ground truth (for both $\mathcal{Y}_{2D}$ and $\mathcal{Y}_{dep}$) and images in the wild with only 2D ground truth (for $\mathcal{Y}_{2D}$). In the reminder of this paper, the 3D and 2D training image sets are denoted as $\mathcal{I}_{3D}$ and $\mathcal{I}_{2D}$, respectively.

\subsection{2D Pose Estimation Module}
\label{sec:2D}

We adopt the state-of-the-art hourglass network architecture in~\cite{newell2016stacked} as our 2D pose estimation module. The network output is a set of $J$ low-resolution heat-maps. Each map $\hat{Y}_{HM} \in \mathcal{R}^{H \times W}$ represents a 2D probability distribution of one joint. The predicted joints in the 2D pose $\hat{Y}_{2D} \in \mathcal{Y}_{2D}$ are the peak locations on these heat-maps. This heat-map representation is convenient as it can be easily combined (concatenate or sum) with the other deep layer feature maps, \eg, as shown in Fig~\ref{fig:Framework}.

To train this module, the loss function is
\begin{equation}
L_{2D}(\hat{Y}_{HM}, Y_{2D}) = \sum_h^H \sum_w^W (\hat{Y}_{HM}^{(h,w)} - G(Y_{2D})^{(h,w)})^2.
\label{Eq:1}
\end{equation}
The loss measures the $L^2$ distance between the predicted heat-maps $\hat{Y}_{HM}$ and the heat-maps $G(Y_{2D})$ rendered from the ground truth $Y_{2D}$ through a Gaussian kernel~\cite{newell2016stacked}.

\subsection{Depth Regression Module}
\label{sec:lifting}

Compared with previous methods that recover 3D joint locations from only 2D joint predictions~\cite{ramakrishna2012reconstructing,zhou20153d,akhter2015pose},
our approach innovates in terms of (i) the integration of 2D and 3D modules for end-to-end network training, and (ii) the usage of a 3D geometric constraint induced loss. They are elaborated below.

\textbf{Integration of 2D and 3D modules.} A key issue for depth estimation is how to effectively exploit image features. A widely used strategy in previous~\cite{zhou2016sparseness,tome2017lifting,chen20163d} is to take the 2D joint locations as the only input for depth prediction as in this way the Mocap-only data can be utilized. However, this strategy is inherently ambiguous, as there typically exist multiple 3D interpretations of a single 2D skeleton. We propose to combine the 2D joint heat-maps and the intermediate feature representations in the 2D module as input to the depth regression module. These features, which extract semantic information at multiple levels for 2D pose estimation, provide additional cues for 3D pose recovery. This shared common feature learning is crucial in our approach.

\textbf{3D geometric constraint induced loss.} One challenge for depth learning is to how to integrate both fully-labeled and weakly-labeled images. For fully-annotated 3D dataset $\mathcal{S}_{3D} = \{\mathcal{I}_{3D}, \mathcal{Y}_{2D}, \mathcal{Y}_{dep}\}$,
the training loss can be simply the standard Euclidean Loss using ground-truth depth label. 
For weakly-labeled dataset $\mathcal{S}_{2D} = \{\mathcal{I}_{2D}, \mathcal{Y}_{2D}\}$, we propose a novel loss induced from a geometric constraint. In the absence of ground truth depth label, this geometric constraint serves as effective regularization for depth prediction.

Overall, let $\hat{Y}_{dep}$ denote the predicted depth. The loss of the depth regression module is 
 \begin{equation}
 \label{Ldep}
 L_{dep}(\hat{Y}_{dep}|I, Y_{2D}) = 
  \left\{
   \begin{array}{c}
   \lambda_{reg}||Y_{dep} - \hat{Y}_{dep}||^2, if \ I \in \mathcal{I}_{3D} \\
   \lambda_{geo}L_{geo}(\hat{Y}_{dep}|Y_{2D}), if \ I \in \mathcal{I}_{2D} \\
   \end{array}
  \right.
\end{equation}
where $\lambda_{reg}$ and $\lambda_{geo}$ are the corresponding loss weights.

$L_{geo}(\hat{Y}_{dep}|{Y}_{2D})$ is the proposed geometric loss. It is based on the fact that ratios between bone lengths remain relative fixed in a human skeleton (e.g., upper/lower arms have a fixed length ratio, left/right shoulder bones share the same length).

Specifically, let $R_i$ be a set of involved bones in a skeleton group $i$, e.g. $R_{arm} = $\{left upper arm, left lower arm, right upper arm, right lower arm\}, let $\l_e$ be the length of bone $e$, and let $\overline{l}_e$ denote the length of bone $e$ in a canonical skeleton (in our experiments, it is set as the average of all training subjects of Human 3.6M dataset). The ratio $\frac{\l_e}{\overline{l}_{e}}$ for each bone $e$ in each group $R_i$ should remain fixed. The proposed loss measures the sum of variance among
$\{\frac{l_e}{\overline{l}_e}\}_{e \in R_i}$ 
of each $R_i$:
\begin{equation}
\label{var}
L_{geo}(\hat{Y}_{dep}|Y_{2D}) = \sum_{i} \frac{1}{|R_i|}\sum_{e \in R_i}{\big(\frac{l_e}{\overline{l}_e}  - \overline{r}_i\big)^2}, 
\end{equation}
where
$$
\overline{r}_i = \frac{1}{|R_i|} \sum\limits_{e \in R_i}\frac{l_e}{\overline{l}_e}.
$$
Note that the bone length is a function of joint locations, which are in turn functions of the predicted depths. Thus, $L_{geo}$ is continuous and differentiable with respect to $\hat{Y}_{dep}$.
The math details of forward and backward equations are provided in the supplemental material
Also note that $L_{geo}$ is defined on the ground truth 2D position $Y_{2D}$ instead of the predicted 2D position $\hat{Y}_{2D}$. This makes the training easier as there is no back-propagation into the 2D module.

In our experiments, we consider $4$ groups of bones:
$R_{arm} = $ \{left/right lower/upper arms\}, $R_{leg} = $ \{ left/right lower/upper legs\}, $R_{shoulder} = $ \{ left/right shoulder bones \}, $R_{hip}$ = \{left/right hip bones\}.
We do not include bones on the torso as we found them exhibit relatively high variance in bone lengths across different human shapes, which makes our constraint less valid.
Note that bones in different sets do not affect each other. 

\subsection{Training}
\label{sec:training}

Combining the losses in Eq.~(\ref{Eq:1}), (\ref{Ldep}), and (\ref{var}), the overall loss for each training image $I \subset \mathcal{I}_{2D}\cup \mathcal{I}_{3D}$ is

\begin{equation}
\begin{split}
L(\hat{Y}_{HM}, \hat{Y}_{dep}|I) = & L_{2D}(\hat{Y}_{HM}, Y_{2D}) + \\
                                   & L_{dep}(\hat{Y}_{dep}|I, Y_{2D}).
\end{split}
\end{equation}

Stochastic gradient descent optimization is used for training. Similar to~\cite{tzeng2015simultaneous} and ~\cite{hoffman2016fcns}, each mini-batch contains both the 2D and 3D training examples (half-half), which are randomly sampled.

In experiments, we found the direct end-to-end training of the whole network from scratch does not work well, likely because of the dependency between the two modules and highly non-linear property of the new geometric constraint induced loss. Thus, we propose a three-stage training scheme that we found is more stable and effective in practice. Note that the final stage is end-to-end.

\emph{Stage 1} initializes the 2D pose module using 2D annotated images, as described in ~\cite{newell2016stacked}. \emph{Stage 2} initializes the 3D pose estimation module and fine-tunes the 2D pose estimation module. Both 2D and 3D annotated data are used. The geometric constraint is not activated, by setting $\lambda_{geo} = 0$ in Equation~\ref{Ldep}. \emph{Stage 3} fine-tunes the whole network with all data. The geometric constraint is activated.

\section{Experimental Evaluation}

To validate our approach, a single model is trained using Human3.6M data~\cite{h36m_pami} and MPII data~\cite{andriluka14cvpr}. Evaluation is performed on three different testing datasets.

The evaluations are from two aspects: supervised 3D human pose estimation (Section~\ref{Sec:Human:3.6M}) and transferred 3D human pose estimation in the wild(Section~\ref{Sec:Inthewild}).

Qualitative results are summarized in Table.~\ref{table:demo}. More qualitative results on MPII validation set can be found in the supplementary material.

\subsection{Experimental Setup}
\label{setup}

\subsubsection{Implementation Detail}

Our method was implemented with torch7~\cite{torch}. 
The hourglass component was based on the public code in~\cite{newell2016stacked}.
For fast training, we used a shallow version of stacked hourglass, i.e. $2$ stacks with $2$ residual modules~\cite{he2016deep} for each hourglass. The depth regression module contains $4$ sequential residual \& pooling modules, which can be regarded as a half hourglass.
The same network architecture and training iterations are used in all of our experiments.

The first training stage in Section ~\ref{sec:training} took $240k$ with a batchsize of $6$. This gave us a 2D pose estimation module with similar performance as in~\cite{newell2016stacked}. 
Stage 2 and stage 3 took $200k$ and $40k$ iterations, respectively. 
The whole training procedure took about two days in one Titan X GPU with CUDA 8.0 and cudnn 5.
A forward pass at testing is about $30ms$.
We set $\lambda_{reg} = 0.1$ and $\lambda_{geo} = 0.01$. We followed~\cite{newell2016stacked} to set all the other hyper-parameters.

\setlength{\tabcolsep}{2pt}
\begin{table*}
\begin{center}
\begin{tabular}{lcccccccc}
\hline\noalign{\smallskip}
 & Directions & Discussion & Eating & Greeting & Phoning & Photo & Posing & Purchases\\
\noalign{\smallskip}
\hline
\noalign{\smallskip}
Chen \& Ramanan~\cite{chen20163d} & 89.87 & 97.57 & 89.98 &  107.87 &  107.31 & 139.17 &  93.56 &  136.09 \\
Tome et al.~\cite{tome2017lifting} & 64.98 & 73.47 & 76.82 & 86.43 & 86.28 & 110.67 & 68.93 & 74.79 \\
Zhou et al.~\cite{zhou2017monocap} &  87.36 & 109.31 &  87.05 & 103.16 & 116.18 & 143.32 & 106.88 & 99.78\\
Metha et al.~\cite{mehta2016monocular} &  59.69 & 69.74 & 60.55 & 68.77 & 76.36 & 85.42 & 59.05 & 75.04\\
Pavlakos et al.~\cite{pavlakos2016coarse} & 58.55 & 64.56 & 63.66 & \textbf{62.43} & 66.93 & 70.74 & 57.72 & 62.51\\
\noalign{\smallskip}
\hline
\noalign{\smallskip}
3D/wo geo  &  73.25 & 79.17 & 72.35 & 83.90 & 80.25 & 81.86 & 69.77 & 72.74 \\
3D/w geo & 72.29 & 77.15 & 72.60 & 81.08 & 80.81 & 77.38 & 68.30 & 72.85 \\
3D+2D/wo geo & 55.17 & 61.16 &  \textbf{58.12} & 71.75 & 62.54 & 67.29 & 54.81 & 56.38 \\
3D+2D/w geo &  \textbf{54.82} & \textbf{60.70} &  58.22 & 71.41 & \textbf{62.03} & \textbf{65.53}  & \textbf{53.83} & \textbf{55.58} \\
\hline\noalign{\smallskip}
 & Sitting & SittingDown & Smoking & Waiting & WalkDog & Walking & WalkPair & Average\\
 \noalign{\smallskip}
\hline
\noalign{\smallskip}
Chen \& Ramanan~\cite{chen20163d} & 133.14 & 240.12 & 106.65 &  106.21 &  87.03 & 114.05 &  90.55 &  114.18 \\
Tome et al.~\cite{tome2017lifting} & 110.19 & 172.91 & 84.95 & 85.78 & 86.26 & 71.36 & 73.14 & 88.39 \\
Zhou et al.~\cite{zhou2017monocap} & 124.52 & 199.23 & 107.42 & 118.09 & 114.23 & 79.39 & 97.70 & 79.9\\
Metha et al.~\cite{mehta2016monocular} & 96.19 & 122.92 & 70.82 & 68.45 & 54.41 & 82.03 & 59.79 &  74.14 \\
Pavlakos et al.~\cite{pavlakos2016coarse} & 76.84 & \textbf{103.48} & 65.73 & \textbf{61.56} & 67.55 & \textbf{56.38} & 59.47 &  66.92\\
\noalign{\smallskip}
\hline
\noalign{\smallskip}
3D/wo geo & 98.41 & 141.60 & 80.01 & 86.31  & 61.89 & 76.32 & 71.47 &  82.44\\
3D/w geo & 93.52 & 131.75 & 79.61 & 85.10  & 67.49 & 76.95 & 71.99 &  80.98\\
3D+2D/wo geo& \textbf{74.79} & 113.99 & 64.34 & 68.78 & 52.22 & 63.97 & 57.31 & 65.69\\
3D+2D/w geo & 75.20 &  111.59 & \textbf{64.15} & 66.05 & \textbf{51.43} & 63.22 & \textbf{55.33} & \textbf{64.90}\\
\hline
\end{tabular}
\caption{Results of Human3.6M Dataset. The numbers are mean Euclidean distance(mm) between the ground-truth 3D joints and the estimations of different methods.}
\label{table:H36M}
\end{center}
\end{table*}
\setlength{\tabcolsep}{1.4pt}

\setlength{\tabcolsep}{2pt}
\begin{table}
\begin{center}
\begin{tabular}{|c|c|c|c|}
\hline
3D/wo geo & 3D/w geo  & 3D+2D/wo geo & 3D+2D/w geo\\
\hline
90.01\% & 90.57\% & 90.93\% & 91.62\%\\
\hline
\end{tabular}
\caption{2D pose accuracy (PCKh@0.5) on Human 3.6M dataset.}
\label{table:H36M2D}
\end{center}
\end{table}
\setlength{\tabcolsep}{1.4pt}

\subsubsection{Datasets \& Metrics}
\label{metric}

\textbf{MPII-training.} MPII dataset ~\cite{andriluka14cvpr} is used for training. It is a large scale in-the-wild human pose dataset.
The images are collected from on-line videos and annotated by human for $J = 16$ 2D joints. 
It contains 25k training images 
and 2957 validation images~\cite{tompson2015efficient}. The human subjects are annotated with bounding boxes. 
We use the training set of MPII to train the 2D pose estimation module. It also provides weak supervision for the depth regression module. 

\textbf{Human3.6M.} Human 3.6M dataset~\cite{h36m_pami} is used both in training and testing. 
It is a widely used dataset for 3D human pose estimation. 
This dataset contains 3.6 millions of RGB images captured by a MoCap System in an indoor environment. 
We down-sampled the video from $50fps$ to $10fps$ for both the training and testing sets to reduce redundancy. Following the standard protocol in ~\cite{li20143d,zhou2016sparseness,zhou2016deep}, we use $5$ subjects(S1, S5, S6, S7, S8) for training and the rest $2$ subjects(S9, S11) for testing. 
The evaluation metric is mean per joint position error(MPJPE) in mm after aligning the depths of the root joints. We use its projected 2D locations for training the 2D module and its depth annotation for depth regression module.

We use the ground truth 2D joint locations provided in the dataset in training (thus implicitly use the camera calibration information), for aligning the 3D and 2D poses. 
During testing, such calibration is not needed, by requiring that the sum of all 3D bones lengths is equal to that of a pre-defined canonical skeleton, as is done in ~\cite{pavlakos2016coarse,zhou2017monocap}. The converting formulation is as follows:
$$
\hat{Y} = (Y_{out} - Y_{out}^{(root)}) * \frac{AvgSumLen}{SumLen_{out}} + Y_{GT}^{(root)}
$$
Where $Y_{out}$ is the combined 2D and depth 3D joint, which is the output of the network; $SumLen_{out}$ is the calculated sum-of-skeleton-length of the output joints;
and $AvgSumLen$ is an constant, which is calculated as the average sum-of-skeleton-length of all the training subjects in Human 3.6M dataset.

\textbf{MPI-INF-3DHP.}
MPI-INF-3DHP~\cite{mehta2016monocular} is a newly proposed 3D human pose dataset. The images were captured by a MoCap system both in indoor and outdoor scenes. We only use its test set split for evaluation. The test set contains $2929$ valid frames from $6$ subjects, performing $7$ actions. 
Following ~\cite{mehta2016monocular}, we employ average PCK (with a threshold $150mm$) and AUC as the evaluation metrics, i.e., after aligning the root joint (pelvis). Note that we assume the global scale is known for experimental evaluation.
We observe that the definition of pelvis position in MPI-INF-3DHP is different from the one used in our training sets (i.e., Human 3.6M and MPII), so we moved the pelvis and hips towards neck in a fixed ratio ($0.2$) as post processing in our evaluation.

\textbf{MPII-Validation.}
Although MPII dataset does not provide 3D pose annotation, we use its validation subset ~\cite{tompson2015efficient}  in our evaluation for two purposes. It contains $2958$ in-the-wild images out of the training set. 

First, we provide qualitative 3D pose estimation results. Many of them looks plausible and convincing. See more in supplementary material. 

Second, we can still evaluate the geometric validity of the estimated 3D pose, which is improved by our proposed constraint. We use the symmetric bone lengths' difference (e.g., left and right upper arms) as the evaluation metric. 
To compute the metric, we normalize the 2D joints in $256 \times 256$ pixels (so that the predicted joints can be directly plotted in the input image). 
The depth is normalized by the same scale.
We then compute the L1 distance between the left and right symmetric bones, e.g. for upper arms it is $||Y^{(left \ shoulder)} - Y^{(left \ elbow)}|| - ||Y^{(right \ shoulder)} - Y^{(right \ elbow)}|| |$. 
This metric is applied for both MPI-INF-3DHP dataset and MPII-Validation set to evaluate the effectiveness of our proposed weakly-supervised geometric loss.

\setlength{\tabcolsep}{2pt}
\begin{table*}
\begin{center}
\begin{tabular}{lccccccccc}
\hline\noalign{\smallskip}
 & Studio GS & Studio no GS & Outdoor & ALL PCK & AUC\\
\noalign{\smallskip}
\hline
\noalign{\smallskip}
Metha et al.(H36M+MPII)~\cite{mehta2016monocular} & 70.8 & 62.3 & 58.8 & 64.7 & 31.7\\
{3D/wo geo}  & 34.4 & 40.8 & 13.6 &  31.5 & 18.0 \\
{3D/w geo}  & 45.6 & 45.1 & 14.4  &  37.7 & 20.9 \\
{3D+2D/wo geo}  & 68.8 & 61.2 & 67.5  & 65.8 & 32.1\\
{3D+2D/w geo}  & 71.1 & 64.7 & 72.7 & \textbf{69.2} & \textbf{32.5}\\
\noalign{\smallskip}
\hline
\noalign{\smallskip}
Metha et al.(MPI-INF-3DHP)~\cite{mehta2016monocular} & 84.1 & 68.9 & 59.6 & \textbf{72.5} & \textbf{36.9}\\
\hline
\end{tabular}
\caption{Results of MPI-INF-3DHP Dataset by scene. GS indicates green screen background. The results are shown in PCK and AUC.}
\label{table:MPI-INF-3DHP}
\end{center}
\end{table*}
\setlength{\tabcolsep}{1.4pt}

\setlength{\tabcolsep}{2pt}
\begin{table}
\begin{center}
\begin{tabular}{lcccccccc}
\hline\noalign{\smallskip}
 & 3D+2D/wo geo  & 3D+2D/w geo\\
\noalign{\smallskip}
\hline
\noalign{\smallskip}
Upper arm & 42.4mm & \textbf{37.8}mm \\
Lower arm & 60.4mm & \textbf{50.7}mm \\
Upper leg & 43.5mm & \textbf{43.4}mm \\
Lower leg & 59.4mm & \textbf{47.8}mm \\
\hline
\hline
Upper arm & 6.27px & \textbf{4.80}px \\
Lower arm & 10.11px & \textbf{6.64}px \\
Upper leg & 6.89px & \textbf{4.93}px \\
Lower leg & 8.03px & \textbf{6.22}px \\
\hline
\end{tabular}
\caption{Evaluation of left-right Symmetry of with and without constraint on MPI-INF-3DHP(Up) and MPII-Validation set (Bottom). Results shown in average L1 distance between left and right bone in mm/3D pixels, respectively}
\label{table:Sym}
\end{center}
\end{table}
\setlength{\tabcolsep}{1.4pt}

\subsubsection{Baselines for Ablation Study}

We implemented three baseline methods and trained the baseline models in the same way as for proposed method.

\textbf{3D/wo geo} It only uses 3D labeled data to train the network in \emph{Stage2} and \emph{Stage3} of Sec. ~\ref{sec:training}. The in-the-wild images are not used. Note that the 2D hourglass module is pre-trained on the 2D dataset in \emph{Stage1}.

\textbf{3D/w geo} It adds the geometric constraint induced loss into the first baseline.

\textbf{3D+2D/wo geo} Its only difference from the proposed method is that the geometric constraint is not utilized for 2D labeled data when training the 3D module.

The proposed method is denoted as \textbf{3D+2D/w geo}.

\subsection{Supervised 3D Human Pose Estimation}
\label{Sec:Human:3.6M}

We first report and analyze the performance of our method on Human 3.6M dataset~\cite{h36m_pami}.

\textbf{Baseline comparison.} Table~\ref{table:H36M} compares the proposed approach with the three baselines. The average MPJPE of baseline \textbf{3D/wo geo} is $82.44mm$. This is already comparable to most state-of-the-art methods~\cite{zhou2016deep,tome2017lifting,zhou2017monocap}.
Note that this baseline is similar with Metha et al.~\cite{mehta2016monocular}, which fine-tuned 2D pose network~\cite{insafutdinov2016deepercut} with 3D data for information transfer. 
The difference is that we did not use $1000 \times$ learning rate decay for the transferred layers, which in our case yielded worse performance.

\begin{table*}
     \begin{center}
     \begin{tabular}{cccccccc}
     \includegraphics[width=0.09\textwidth]{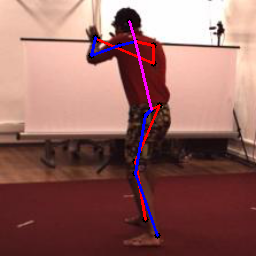}
      &\includegraphics[width=0.13\textwidth]{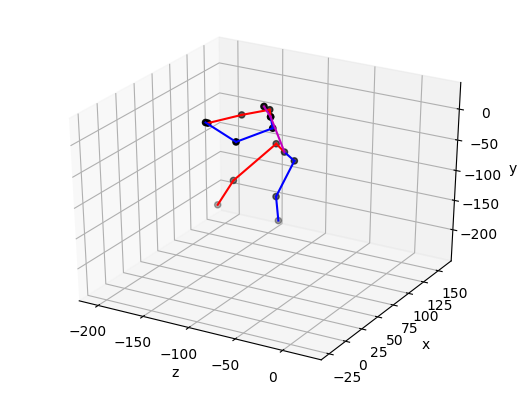}
      &\includegraphics[width=0.09\textwidth]{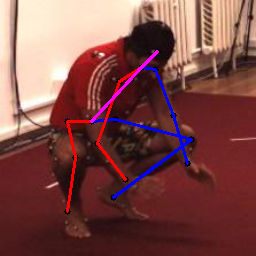}
      &\includegraphics[width=0.13\textwidth]{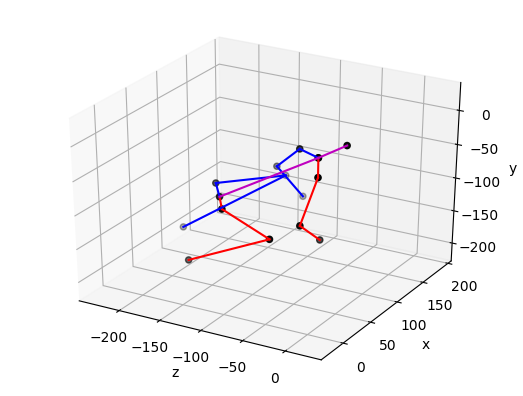}
      &\includegraphics[width=0.09\textwidth]{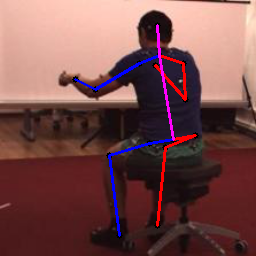}
      &\includegraphics[width=0.13\textwidth]{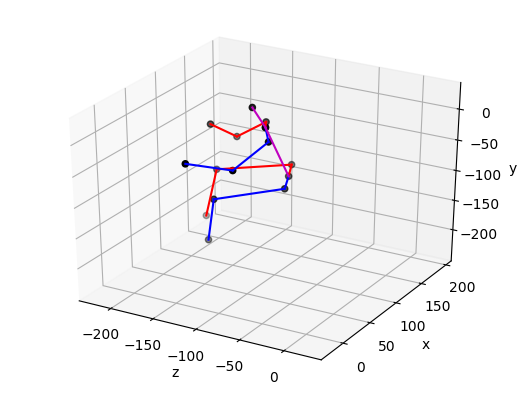}
      &\includegraphics[width=0.09\textwidth]{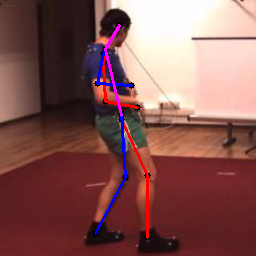}
      &\includegraphics[width=0.13\textwidth]{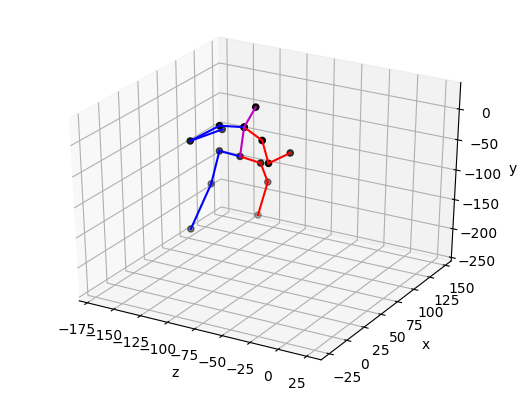} \\ 
     \includegraphics[width=0.09\textwidth]{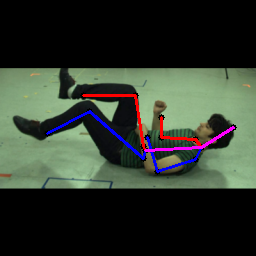}
      &\includegraphics[width=0.13\textwidth]{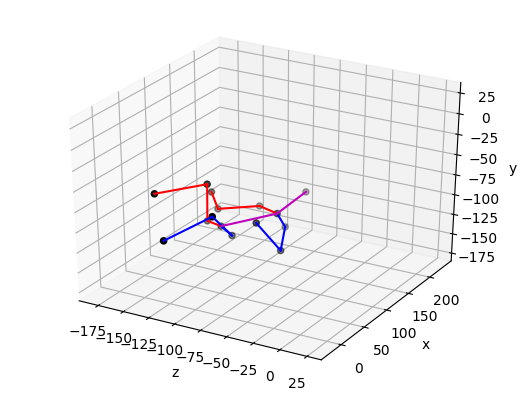}
      &\includegraphics[width=0.09\textwidth]{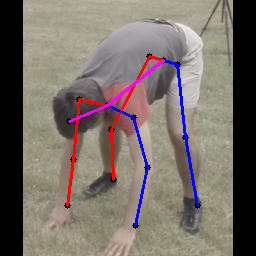}
      &\includegraphics[width=0.13\textwidth]{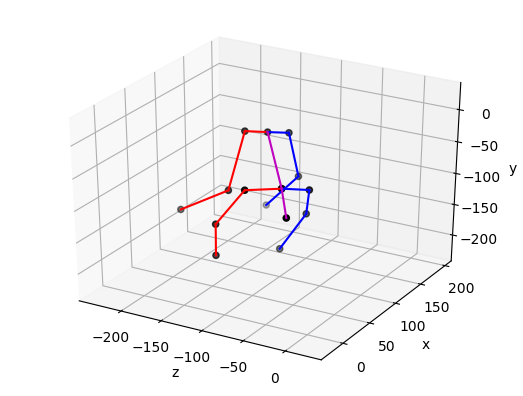}
      &\includegraphics[width=0.09\textwidth]{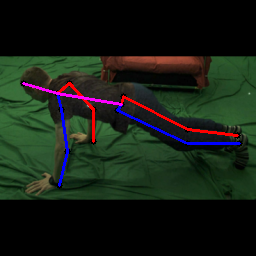}
      &\includegraphics[width=0.13\textwidth]{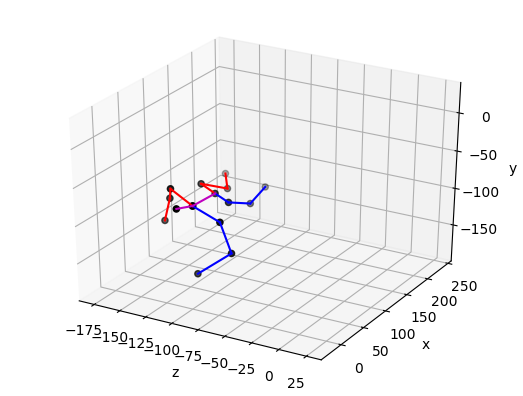}
      &\includegraphics[width=0.09\textwidth]{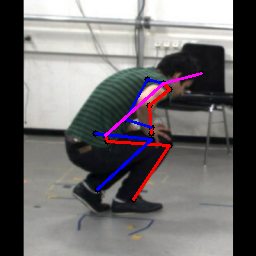}
      &\includegraphics[width=0.13\textwidth]{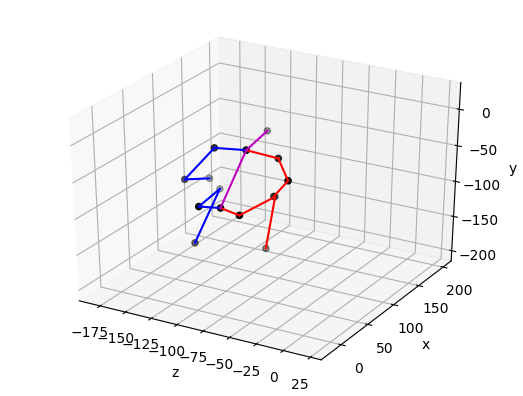} \\
     \includegraphics[width=0.09\textwidth]{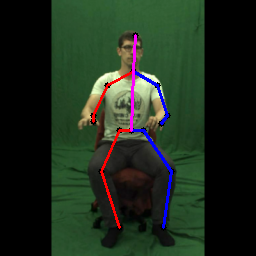}
      &\includegraphics[width=0.13\textwidth]{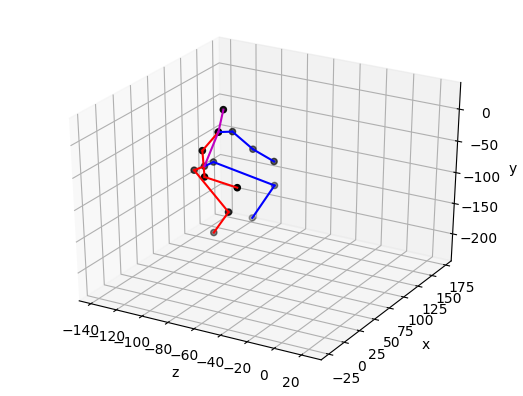}
      &\includegraphics[width=0.09\textwidth]{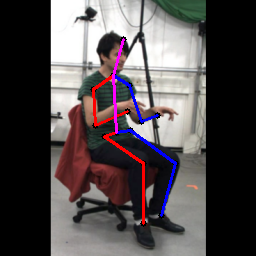}
      &\includegraphics[width=0.13\textwidth]{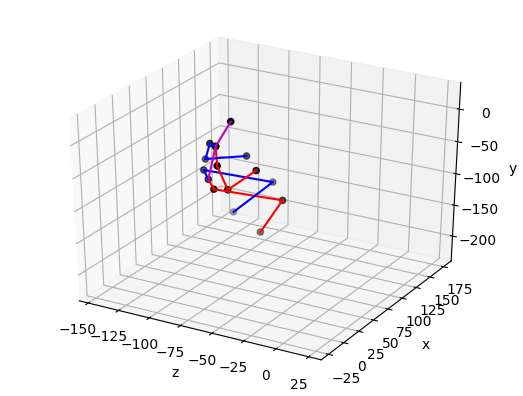}
      &\includegraphics[width=0.09\textwidth]{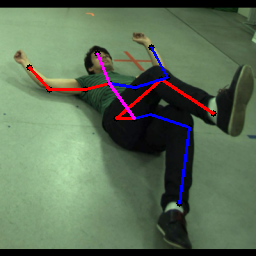}
      &\includegraphics[width=0.13\textwidth]{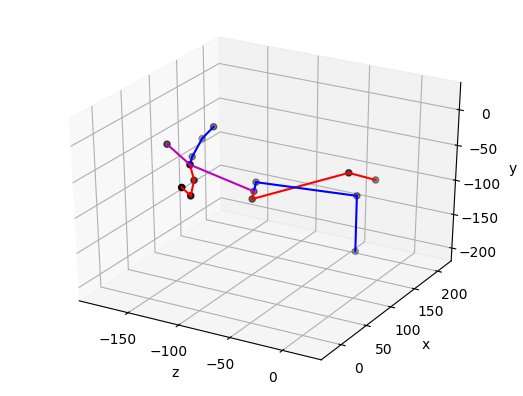}
      &\includegraphics[width=0.09\textwidth]{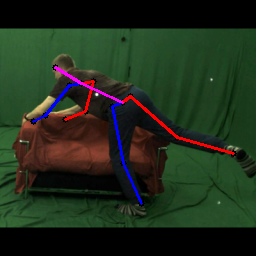}
      &\includegraphics[width=0.13\textwidth]{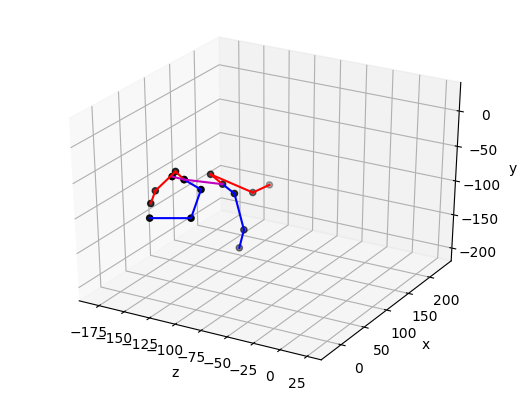} \\
     \includegraphics[width=0.09\textwidth]{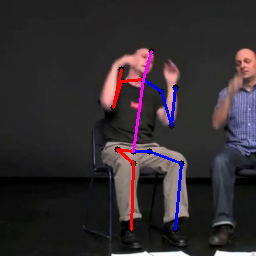}
      &\includegraphics[width=0.13\textwidth]{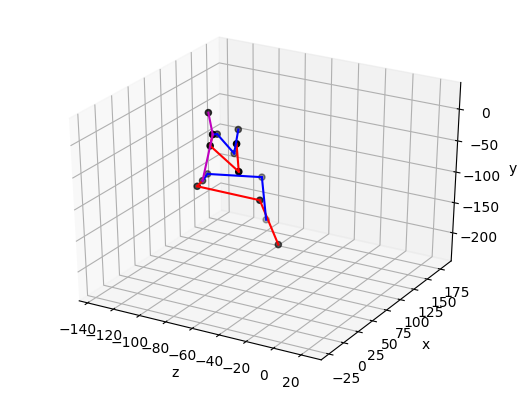}
      &\includegraphics[width=0.09\textwidth]{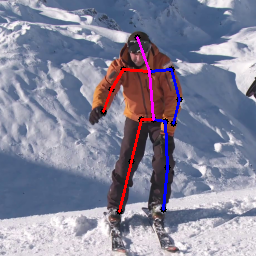}
      &\includegraphics[width=0.13\textwidth]{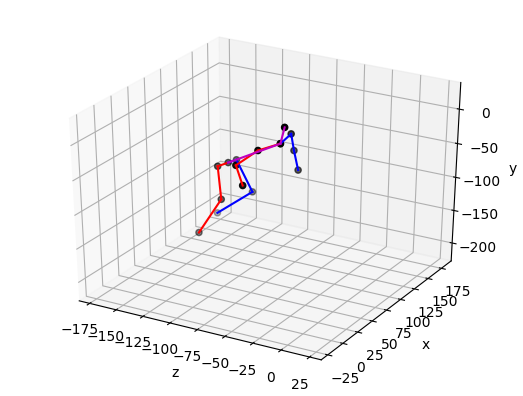}
      &\includegraphics[width=0.09\textwidth]{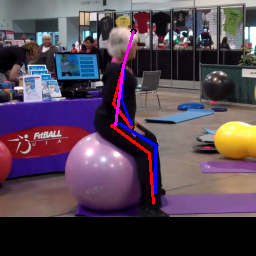}
      &\includegraphics[width=0.13\textwidth]{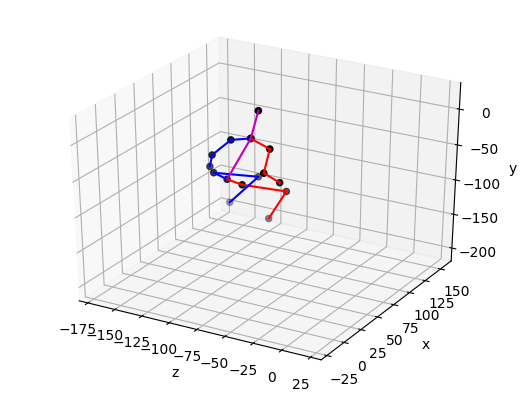}
      &\includegraphics[width=0.09\textwidth]{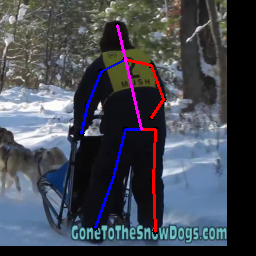}
      &\includegraphics[width=0.13\textwidth]{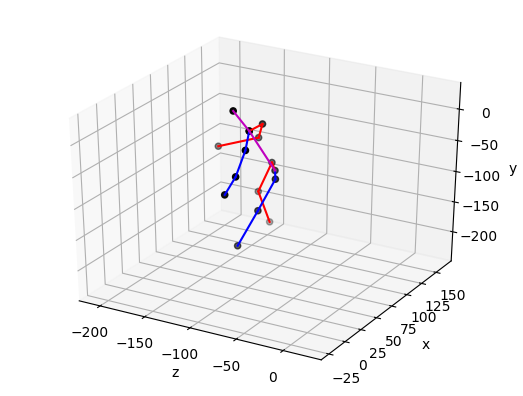} \\
     \includegraphics[width=0.09\textwidth]{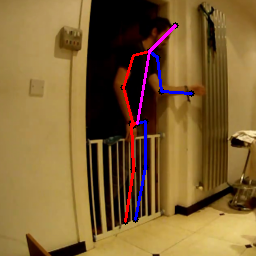}
      &\includegraphics[width=0.13\textwidth]{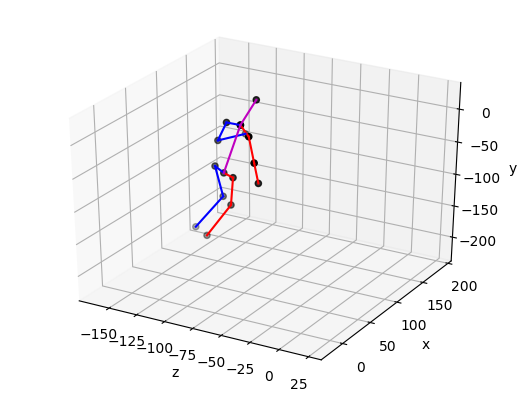}
      &\includegraphics[width=0.09\textwidth]{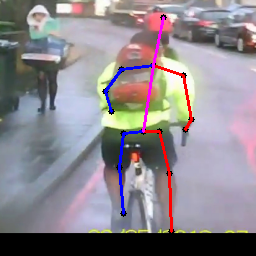}
      &\includegraphics[width=0.13\textwidth]{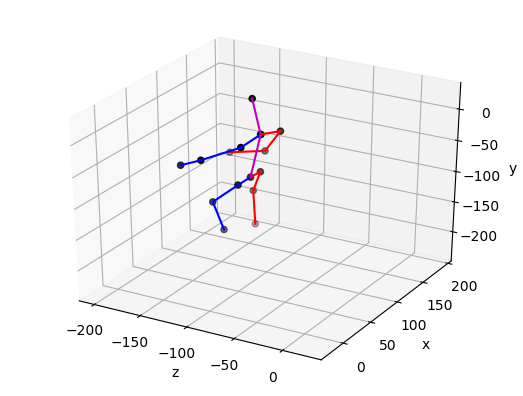}
      &\includegraphics[width=0.09\textwidth]{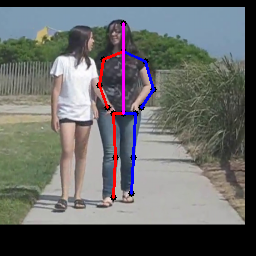}
      &\includegraphics[width=0.13\textwidth]{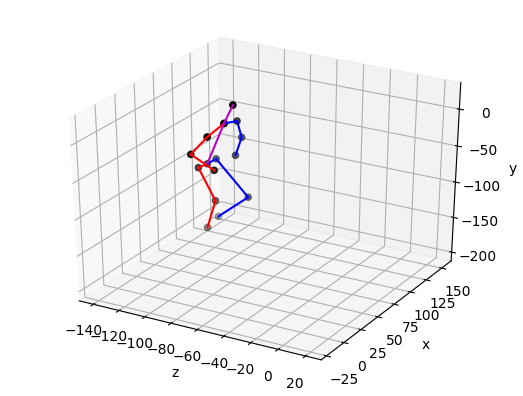}
      &\includegraphics[width=0.09\textwidth]{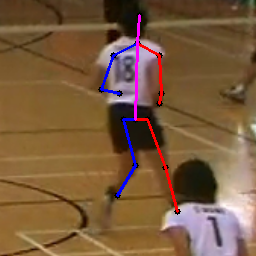}
      &\includegraphics[width=0.13\textwidth]{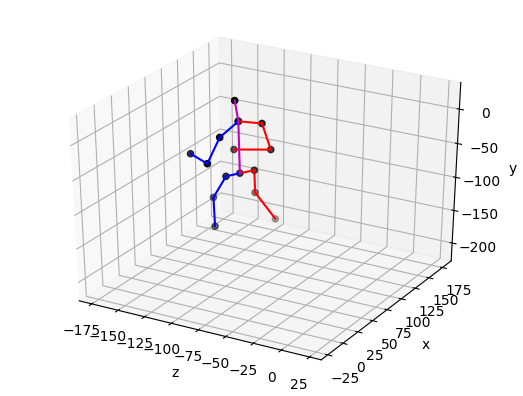} \\
     \includegraphics[width=0.09\textwidth]{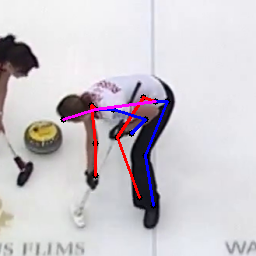}
      &\includegraphics[width=0.13\textwidth]{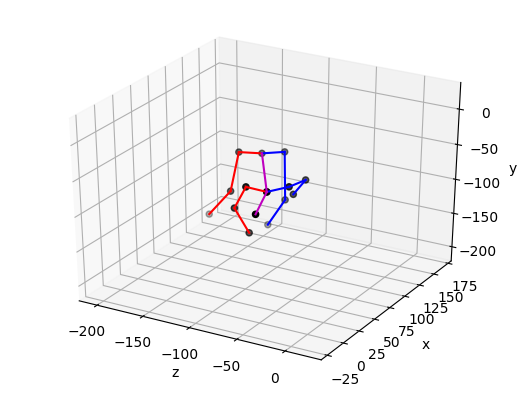}
      &\includegraphics[width=0.09\textwidth]{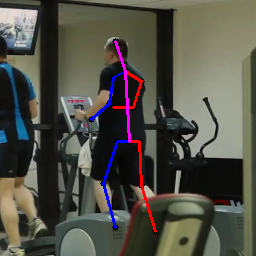}
      &\includegraphics[width=0.13\textwidth]{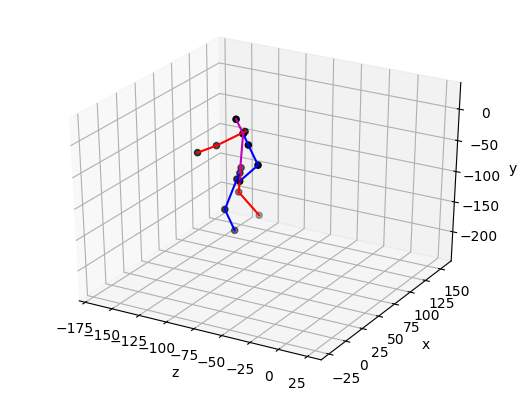}
      &\includegraphics[width=0.09\textwidth]{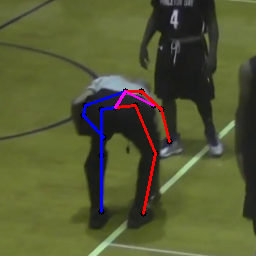}
      &\includegraphics[width=0.13\textwidth]{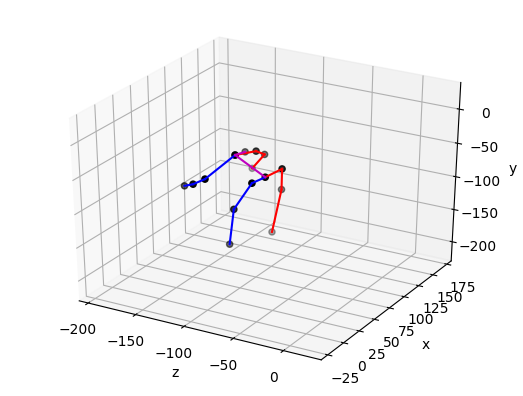}
      &\includegraphics[width=0.09\textwidth]{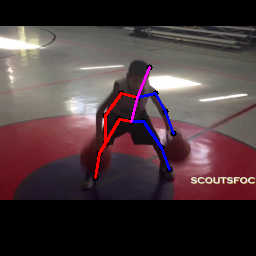}
      &\includegraphics[width=0.13\textwidth]{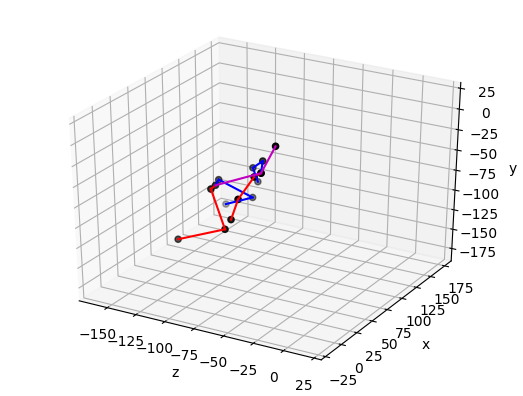} \\
     \includegraphics[width=0.09\textwidth]{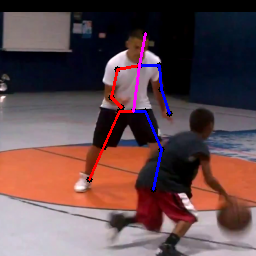}
      &\includegraphics[width=0.13\textwidth]{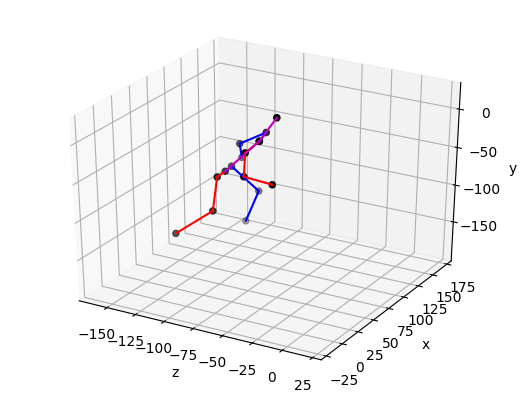}
      &\includegraphics[width=0.09\textwidth]{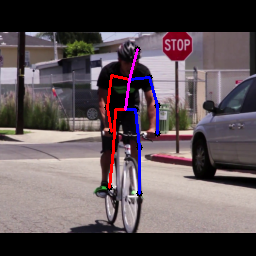}
      &\includegraphics[width=0.13\textwidth]{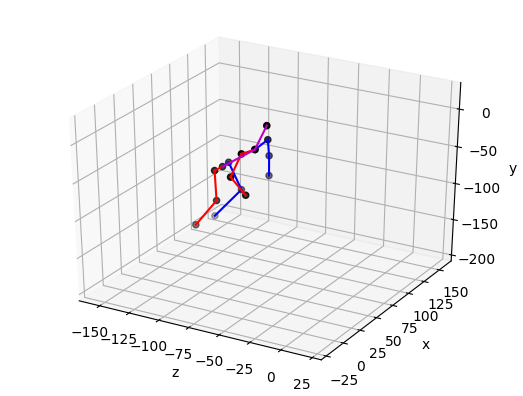}
      &\includegraphics[width=0.09\textwidth]{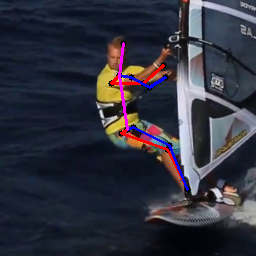}
      &\includegraphics[width=0.13\textwidth]{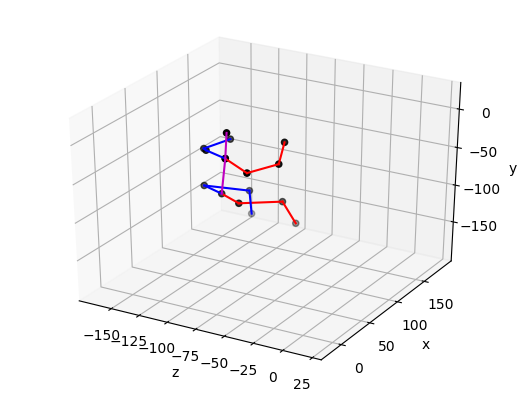}
      &\includegraphics[width=0.09\textwidth]{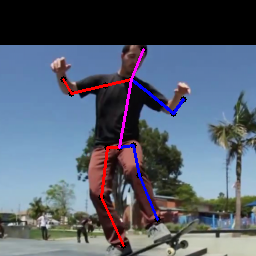}
      &\includegraphics[width=0.13\textwidth]{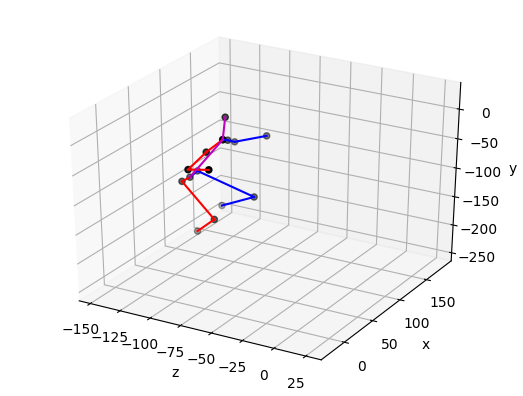} \\
      \end{tabular}
      \caption{Qualitative results from different datasets. We show the 2D pose on the original image and 3D pose from a novel view. First line: Human 3.6M dataset; Second and third lines: MPI-INF-3DHP dataset; Fourth to seventh lines: MPII dataset.}
      \label{table:demo}
      \end{center}
\end{table*}

Adding the geometric constraint, \ie, \textbf{3D/w geo}, provides a decent performance gain.

Training with both 2D and 3D data (\textbf{3D+2D/wo geo}), provides significant performance gain --- average MPJPE dropped to $64.90mm$, which is superior to all previous work~\cite{mehta2016monocular,pavlakos2016coarse}. This verifies the effectiveness of combining data sources in our unified training.

Finally, the proposed approach \textbf{3D+2D/w geo} achieves the best results. Note that the constraints are applied on the disjoint 2D dataset, showing that the provided prior knowledge is universal. 
We have also tested adding constraints on fully-supervised 3D data. 
The results are similar.

\textbf{Comparisons to other \emph{in-the-wild} methods.} 
Our method is superior to other methods that are applicable to in-the-wild images. 
Comparing to two two-step methods,
MPJPE of Chen \& Ramanan~\cite{chen20163d} is $114.18mm$ and MPJPE of Zhou et al.~\cite{zhou2017monocap} is $79.9mm$. 
Pavlakos et al.~\cite{pavlakos2016coarse} provided an alternative decoupled version which can also be applied in the wild, but its MPJPE increased to $78.1mm$. 
MPJPE of our method is $64.90mm$ and significantly better.

\textbf{Why combining 2D and 3D data is better?} A reasonable question is that it is still unclear whether the benefit of combined  training comes from better depth estimation, or just from more accurate 2D pose estimation.

To answer this question, we only evaluate the accuracy of the 2D pose estimation, using the standard metric PCKh@0.5 (see~\cite{andriluka14cvpr}). The results in Tab.~\ref{table:H36M2D} show that the 2D pose is very accurate in all the three baselines and the proposed method. This convincingly indicates that adding 2D data into training \emph{does not} improve the 2D accuracy but mostly benefits the  the depth regression module via shared deep feature representation.

\subsection{Transferred Human Pose In the Wild}
\label{Sec:Inthewild}

We evaluate the generalization of our method on two datasets captured in different in-the-wild environments.

\subsubsection{MPI-INF-3DHP Dataset}
It exhibits considerable domain shift from both MPII and Human 3.6M datasets. 
Table~\ref{table:MPI-INF-3DHP} compares the performance of various methods on MPI-INF-3DHP. 
In this case, the first two baseline methods, i.e., \textbf{3D/wo geo} and \textbf{3D/w geo}, have low performance. This is not surprising, as the 3D training set contains only indoor images. 
We note that even in this case, the geometric constraint is still effective (\textbf{3D/wo geo} is worse than \textbf{3D/w geo}).

\textbf{3D+2D/wo geo} achieved $65.8$ and $32.1$ in PCK and AUC, respectively. 
These numbers are better than their counterparts ($64.7$ PCK and $31.7$ AUC) in ~\cite{mehta2016monocular} 
with Human 3.6M training data, 
again showing the advantage of our training scheme. 

The proposed approach yields $69.2$ in PCK and $32.5$ in AUC.
These numbers are close to the one that is derived from the original training data of MPI-INF-3DHP ~\cite{mehta2016monocular}, which has $72.5$ in PCK and $36.5$ in AUC. 
Our result is strong even though we didn't use their training data.
This confirms the ability of our method on in-the-wild images. 

We also tested the left-right symmetry as described in Sec. ~\ref{metric}. The results in Table. ~\ref{table:Sym} (Bottom) shows that using the geometric constraint considerably improves the geometric validity.

\subsubsection{MPII Validation Dataset}

Finally, we evaluate our method on the most challenging in-the-wild MPII validation set. The qualitative 3D pose results in Table~\ref{table:demo} are quite plausible.

\textbf{Geometric validity.} 
As explained in sec. ~\ref{metric}, we evaluate the left-right symmetry metric. The results in Table~\ref{table:Sym} (Top) show that our approach is considerably better.

\textbf{2D accuracy versus 3D accuracy.} 
We note that our method has a slightly lower 2D joint accuracy than the original Hourglass model. 
This can be expected as our model learns the additional depth regression task.
However, utilizing the geometric constraint improves the 2D joint accuracy as well. 
This indicates that our network is able to propagate this geometric constraint from the 3D module to the 2D module, which justifies the design goal of our network.

\section{Future Work and Conclusions}

In this paper, we introduced an end-to-end system that combines 2D pose labels in the wild and 3D pose labels in restricted environments for the challenge problem of 3D human pose estimation in the wild. 
In the future, we plan to explore more un-/weakly-supervised constraints for a better transfer, e.g., a domain alignment network as in ~\cite{hoffman2016fcns,tzeng2015simultaneous}. 
We hope this work can inspire more works on un-/weakly-supervised transfer learning and on 3D human pose estimation in the wild.
\section*{Acknowledgements}

We thank Dushyant Mehta and Helge Rhodin for helping about evaluating on MPI-INF-3DHP dataset and thank Danlu Chen for help with Fig. ~\ref{fig:Framework}. Also, we thank Wei Zhang for helpful discussion. This work is supported in part by the National Natural Science Foundation of China (\#U1611461, \#61572138), Shanghai Municipal Science and Technology Commission (\#16JC1420401).


{\small
\bibliographystyle{ieee}
\bibliography{egbib}
}

\end{document}